\documentclass[runningheads]{llncs}
\usepackage[T1]{fontenc}
\usepackage{graphicx}
\usepackage{booktabs}

\begin{document} 
\title{Synthetic Data Outliers: Navigating Identity Disclosure}
%
%
\author{Carolina Trindade\inst{1}\orcidID{0009-0007-3950-2533} \and
Luís Antunes\inst{1,2}\orcidID{0000-0002-9988-594X} \and
Tânia Carvalho\inst{1}\orcidID{0000-0002-7700-1955} \and
Nuno Moniz\inst{3}\orcidID{0000-0003-4322-1076}}
\authorrunning{C. Trindade et al.}
%
\institute{Faculdade de Ciências da Universidade do Porto, Porto, Portugal \and TekPrivacy, Porto, Portugal \and Lucy Family Institute for Data \& Society, Indiana, USA}
\maketitle              

\begin{abstract} 
Multiple synthetic data generation models have emerged, among which deep learning models have become the vanguard due to their ability to capture the underlying characteristics of the original data. However, the resemblance of the synthetic to the original data raises important questions on the protection of individuals' privacy. As synthetic data is perceived as a means to fully protect personal information, most current related work disregards the impact of re-identification risk. In particular, limited attention has been given to exploring outliers, despite their privacy relevance. In this work, we analyze the privacy of synthetic data w.r.t the outliers. Our main findings suggest that outliers re-identification via linkage attack is feasible and easily achieved. Furthermore, additional safeguards such as differential privacy can prevent re-identification, albeit at the expense of the data utility.

\end{abstract}

\section{Introduction}\label{sec:intro}
Synthetic data~\cite{rubin1993statistical} refers to data generated by methods designed to capture the distribution and statistical properties of the original dataset, producing data with similar characteristics. In the scope of data protection, synthetic data surged as an alternative to traditional methods such as generalization~\cite{figueira2022survey}. Legal frameworks like the General Data Protection Regulation (GDPR) highlight the need to adopt privacy measures to protect private information. As a result, synthetic data has become prevalent across different industries as a proxy for original data in many tasks, like software testing, where access to high-quality data is challenging due to privacy restrictions~\cite{el2020practical}. Numerous synthetic data generation approaches have been proposed in the literature~\cite{figueira2022survey}, for instance, models based on deep learning, such as Generative Adversarial Networks~\cite{goodfellow2014generative} or Variational Autoencoders~\cite{kingma2013auto}. However, the greater the similarity between the synthetic data and the original data, the greater the risk of disclosure.

Although synthetic data generation is commonly seen as a means to fully protect the privacy of individuals, it has also raised numerous questions about the memorization of deep learning models~\cite{nikolenko2019synthetic}. In particular, outliers are potentially more susceptible to attack due to their deviation from the general data. These extreme data points, often correspond to sensitive information. Existing literature focuses on assessing privacy risks through the application of diverse attack methodologies to quantify different privacy vulnerabilities, such as Membership Inference Attacks (MIA)~\cite{el2020evaluating,houssiau2022tapas,giomi2022unified,stadler2022synthetic}. Stadler et al.~\cite{stadler2022synthetic} also evaluate privacy gain in synthetic data sets using outliers, but only focus on a small portion of rare cases, namely five data points. Although it is theoretically known that extreme data points are particularly more susceptible to intruders' attacks, empirical studies confirming this are scarce.
To the best of our knowledge, no studies have yet focused exclusively on investigating the re-identification risk through linkage attacks of synthetic datasets concerning outliers.

In this paper, we conduct a linkage attack in a group of outliers to demonstrate the effectiveness of re-identification and how easily an attacker can link back supposedly protected personal information. We fine-tuned deep learning and differential privacy-based models to generate multiple synthetic data variants to evaluate their capacity to protect outliers. 

Our main findings are summarized as follows.
\begin{itemize}
    \item The effectiveness of protecting outliers during the synthesis process is ultimately model-dependent;
    \item Differential privacy-based models are more efficient in protecting the privacy of individuals than deep learning-based models, however with higher data utility degradation; and,
    \item Deep learning-based models provide high data quality with increased \textit{epochs}, but at the expense of individuals' privacy.
\end{itemize}

Experiments were run using an Inter Core i7 Processor (4x1.80GHz) and 8GB RAM in a Ubuntu 18 partition with 100GB. Our experimental evaluation focuses on one original dataset due to computational limitations. However, it is noteworthy that \textbf{the re-identification of a single instance renders the dataset subject to the provisions outlined in the GDPR}.

The remainder of this paper is organized as follows. Section~\ref{sec:related} provides an overview of data privacy and current related work. Section~\ref{sec:development} presents the methodology and materials used in our experiments. The results are presented in Section~\ref{sec:results} and discussed in Section~\ref{sec:discussion}. Section~\ref{sec:conclusion} concludes the paper.

\section{Literature Review}\label{sec:related}
In this section, we focus on the main strategies for de-identification, namely, traditional techniques and deep learning-based solutions, including privacy and utility measures. We also discuss our contributions to the related state-of-the-art.

\subsection{Notions}
The de-identification process was designed for secure data publication, aiming to modify data until an acceptable level of disclosure risk and data utility is achieved~\cite{carvalho2023survey}. In this process, quasi-identifiers (QIs) are selected to measure the risk of disclosure. Such attributes, when combined, can provide information that could lead to re-identification of individuals (e.g. gender and date of birth). Privacy-Preserving Techniques (PPTs) are then applied according to the level of privacy and utility. If the desired level is not met, the parameters of the PPTs are adjusted or another PPT is selected. Otherwise, the data is ready for release.

Multiple PPTs were proposed in the literature to reduce disclosure risk while maintaining data utility useful for tasks such as decision-making. Currently, synthetic data is at the forefront of data de-identification methods, given its reputation for providing potentially secure data while maintaining utility. Deep learning-based models have been proven to be more versatile and to provide better results in comparison to other approaches, such as interpolation methods~\cite{figueira2022survey}. Despite their popularity, the generation of synthetic data may overlook the protection of outliers, which makes them more vulnerable to exploitation~\cite{carvalho2023survey}. 

Outliers are data points that deviate from the expected patterns or central tendencies of a dataset~\cite{grubbs1969procedures}. They can be classified as univariate, which are extreme values in the distribution of a particular attribute, or multivariate, which are unique combinations of values in the observations. The most common approach for outlier detection is based on the concept of standard deviation~\cite{mateo2004outlier}. Let $X=\{x_1, x_2,..., x_n\}$ be a dataset with $n$ observations, in which $\mu$ represents the mean and $\sigma$ corresponds to the standard deviation of the dataset. Also, consider a threshold value $k$, that is used to determine the boundary beyond which data points are considered outliers. The outlier detection is then performed by

\begin{definition}[\textbf{Outlier detection}]
Any data point $x_i$ that lies outside the range [$\mu-k\sigma$, $\mu+k\sigma$] is an outlier.
\end{definition}

Researchers have developed a plethora of methods for identifying and handling outliers in datasets. Examples include Tukey's fences~\cite{tukey1977exploratory}, Peirce's criterion~\cite{peirce1852criterion} and graphical approaches such as box plots or normal probability plots. The choice of method generally depends on factors such as the distribution of the data and the desired level of sensitivity to outliers.

Concerning the privacy evaluation, we focus on identity disclosure, also known as re-identification, which occurs when an attacker associates a record of the released dataset as belonging to an individual w.r.t QI values. This type of disclosure can occur under a linkage threat~\cite{article2014opinion}. Common approaches for record linkage are based on the similarity functions between pairs of records~\cite{pagliuca1999some,fellegi1969theory,muralidhar2016rank}.

Many privacy measures have been proposed in the literature. The most known is $k$-anonymity~\cite{samarati2001protecting}, used to evaluate identity disclosure by indicating whether a dataset respects the desired level of $k$. Each individual cannot be distinguished from at least $k$-1 other individuals. A more sophisticated measure is Differential Privacy (DP)~\cite{dwork2006differential} which provides a rigorous mathematical definition of privacy guarantees. With DP, the disclosure risk is assessed avoiding assumptions on previous background knowledge that an attacker may have or may learn about individuals. This is achieved by bounding the sensitivity of released data to the presence of any individual within the dataset. Generally, the Laplace distribution is used to add noise to the output of $K$, which is called Laplace Mechanism.

\begin{definition}[\textbf{Differential Privacy}~\cite{dwork2006differential}]
A randomized function K is \sloppy ${\epsilon}$-differential privacy if for all data sets $D{_1}$ and $D{_2}$ differing on at most one element, and all S ${\subseteq}$ Range(K), $P{_r}|K(D_1) \in S| \leq exp(\epsilon) \times P{_r}|K(D_2) \in S|$. 
\end{definition}

A crucial aspect of de-identification is also the evaluation of utility since the transformations applied can potentially reduce the utility of the data. Thus, utility measures are used to quantify the similarity between the distributions of the original and the de-identified data~\cite{carvalho2023survey}. Information loss measures aim to compare records between the original and the de-identified datasets as well as statistics computed from both to assess if the transformed data is still analytically valid (e.g. discernibility~\cite{bayardo2005data}). Measures such as mean, variance, and correlation are used to quantify the changes in statistics. Predictive performance metrics evaluate the ability to make predictions (e.g. Accuracy, Precision and Recall~\cite{allen1955machine}).

\subsection{Current research directions}
Tao et al.~\cite{tao2021benchmarking} benchmark differentially private synthetic data generation algorithms and classify them as GAN-, Marginal- and Workload-based. Conclusions indicate that Marginal-based methods outperform the remaining approaches and GAN-based methods could not preserve the 1-dimensional statistics of data. Hotz et al.~\cite{hotz2022balancing} discuss the major disadvantages of the use of synthetic data with differential privacy. The authors suggest that, the more accurate the synthesis process in representing the original data, which may contain outliers, the greater the risk of disclosure, thus synthetic data could be used by an attacker to make confident guesses on the identity or attributes of a real person.

El Emam et al.~\cite{el2020evaluating} propose a full risk model to evaluate \sloppy{re-identification} and an attacker's ability to learn new information about an individual from matches between the original and synthetic records. Their findings indicate a low identity disclosure risk in both cases and concluded that the use of synthetic data reduces re-identification significantly. Houssiau et al.~\cite{houssiau2022tapas} propose TAPAS: a Toolbox for Adversarial Privacy Auditing of Synthetic Data. This threat modeling framework defines privacy attacks considering different attacker background knowledge and their learning objectives to evaluate synthetic data, additionally providing analysis reports. Giomi et al.~\cite{giomi2022unified} present Anonymeter, a statistical framework aiming to perform attack-based evaluations of different types of privacy risks in synthetic tabular datasets. In their experiments, synthetic data has the lowest vulnerability against linkability, suggesting that one-to-one relationships between real and synthetic data records are not preserved. Stadler et al.~\cite{stadler2022synthetic} propose a framework that allows the privacy evaluation of differentially private synthetic data and compares it to standard techniques. The authors conclude that synthetic data generated with generative models without any layer of protection do not protect outliers from linkage attacks and differentially private synthetic data protect against membership inference with a large loss of utility.

Previous studies have explored the privacy of synthetic data by conducting diverse attacks to measure different types of privacy risk~\cite{el2020evaluating,houssiau2022tapas,giomi2022unified,stadler2022synthetic}. Among these, the work by 
Stadler et al.~\cite{stadler2022synthetic} is most closely aligned with ours; the authors assess the privacy gain of synthetic data records, including outliers, with and without differential privacy regarding linkage attacks. While the authors perform a complex game to verify privacy gain, our approach involves a simpler linkage attack to demonstrate how easy is to perform a re-identification on outlier records. Furthermore, in contrast to their work, which focuses on only five extreme data points, our analysis includes a dataset with a substantially higher portion of outliers. Additionally, we generate multiple synthetic data variants using three deep learning models and three differential privacy models, each adjusted with an extensive range of hyperparameters, to analyze the influence of specific hyperparameters in synthetic data quality and privacy protection.

\section{Experimental Evaluation}\label{sec:development}
In this section, we provide our experimental methodology by describing the data and methods used. Given an original dataset, we generate multiple synthetic data variants using deep learning and differential privacy models. Then, we evaluate the generated synthetic variants in terms of privacy and utility. Concerning data utility, we use different metrics to analyze how synthetic data retains the original statistical properties. For privacy, we perform a linkage attack by comparing each data variant with the original data w.r.t a set of QIs in the outliers subset.

\subsection{Data}
We use the Credit Risk dataset~\cite{credit} as the original dataset for our study. This dataset has 22.910 records and presents a critical portion of outliers, essential for our experimental evaluation. Table~\ref{tab:attrs} contains a summary of the attributes.

\begin{table*}
\centering
\scriptsize
  \caption{Summary of the attributes of the Credit Risk dataset.}
  \label{tab:attrs}
  \begin{tabular}{lll}
    \toprule
    \textbf{Attribute} & \textbf{Type} & \textbf{Description}\\
    \midrule
    {\itshape person\_age} & Numerical & The person's age\\
    {\itshape person\_income} & Numerical & The person's annual income\\
    {\itshape person\_home\_ownership} & Categorical & The person's home ownership\\
    {\itshape person\_emp\_length} & Numerical & The person's employment length in years\\
    {\itshape loan\_intent} & Categorical & The loan intent\\
    {\itshape loan\_grade} & Categorical & The loan grade\\
    {\itshape loan\_amnt} & Numerical & The loan amount\\
    {\itshape loan\_int\_rate} & Numerical & The loan interest rate\\
    {\itshape loan\_status} & Numerical & The loan status\\
    {\itshape loan\_percent\_income} & Numerical & The loan percent income\\
    {\itshape cb\_person\_default\_on\_file} & Categorical & The historical default\\
    {\itshape cb\_person\_cred\_hist\_length} & Numerical & The credit history length\\
  \bottomrule
\end{tabular}
\end{table*}

This dataset has approximately 12\% missing records. To preserve the integrity of future analysis, all instances of missing values have been removed.

Considering the data properties, we select {\itshape person\_age}, {\itshape person\_income}, {\itshape person\_home\_ownership} and \sloppy {\itshape loan\_intent} as QIs. We hypothesize that their accessibility could facilitate easier re-identification by an attacker.

\subsection{Methods}
\subsubsection{Synthetic Data Generation.}
We use models available in the SDV~\cite{patki2016synthetic,sdv} and DPART~\cite{mahiou2022dpart,dpart} tools to obtain respectively both deep learning- and differential privacy-based synthetic dataset variants. Regarding SDV, we use the TVAE, CTGAN and CopulaGAN models. From DPART, we use the Independent, PrivBayes and DPsynthpop models. The combination of the hyperparameters resulted in 27 synthetic variants for each SDV model and 7 synthetic variants for each DPART model, all with the same size as the original dataset. Thus, in total, we generated 102 synthetic dataset variants. Table~\ref{tab:pars} contains a summary of the hyperparameters tested in this experiment.

\begin{table*}
\centering
\scriptsize
  \caption{Synthetic data generation tools with models and hyperparameters used.}
  \label{tab:pars}
  \begin{tabular}{lll}
    \toprule
    \textbf{Tool} & \textbf{Models} & \textbf{Parameters}\\
    \midrule
    & TVAE & {\itshape epochs} $ \in \{150, 300, 500\}$\\
    SDV & CTGAN & {\itshape batch\_size} $ \in \{20, 50, 100\}$\\
    & CopulaGAN & {\itshape embedding\_dim} $ \in \{12, 64, 128\}$\\
    \hline
    & Independent &\\
    DPART & PrivBayes & {\itshape epsilon}   $ \in \{0.01, 0.1, 0.2, 0.5, 1.0, 5.0, 10.0\}$\\
    & DPsynthpop & \\
  \bottomrule
\end{tabular}
\end{table*}

\vspace{-2em}

\subsubsection{Data Utility.}
In terms of data utility, we use the SDMetrics~\cite{sdmetrics} tool which is integrated with SDV and provides various metrics to evaluate different aspects of synthetic data. We need both original and synthetic dataset variants for such an evaluation. Table~\ref{tab:metrics} contains a summary of the used metrics.

\begin{table}
\centering
\scriptsize
  \caption{Utility metrics used from SDMetrics~\cite{sdmetrics}.}
  \label{tab:metrics}
  \begin{tabular}{p{0.25\linewidth} p{0.55\linewidth}}
    \toprule
    \textbf{Metric} & \textbf{Description}\\
    \midrule
    BoundaryAdherence & Verifies whether a synthetic column respects the minimum and maximum values of the real column.\\
    CategoryCoverage & Verifies whether a synthetic column covers all the possible categories that are present in a real column.\\
    RangeCoverage & Verifies whether a synthetic column covers the full range of values that are present in a real column.\\
    StatisticSimilarity & Verifies the similarity between a real and a synthetic column by comparing a summary statistic (median).\\
  \bottomrule
\end{tabular}
\end{table}

CategoryCoverage and RangeCoverage are equivalent metrics for categorical and numerical attributes, so we combined them as AttributeCoverage.

\subsubsection{Outliers Attack.}
As we conduct linkage attacks on outliers, we use the $z-score$ method to select the outliers subset. This method is described as follows.
\begin{equation}
  Z = \frac{x_i-\mu}{\sigma}
\end{equation}
$Z-score$ is applied to numerical QIs, where $x_i$ represents the $i^{th}$ the attribute value, $\mu$ is the sample mean and $\sigma$ is the standard deviation. We set a threshold $k = 3$; thus, $x_i$ is an outlier if the $z-score$ is greater than 3 or less than -3. As we have multiple numerical QIs, we perform the multivariate outliers selection.

To perform a linkage attack on outliers we use Record Linkage~\cite{rl} toolkit. The parameters in Record Linkage depend on QIs type. For the numerical QIs we used the Gauss method with {\itshape offset} and {\itshape scale} of 5 for the {\itshape person\_age} attribute and 1.000 for the {\itshape person\_income} attribute. These parameters allow us to have a wider range of values to consider as potential matches. For the categorical attributes, we used the Levenshtein method without defining a range, because the package looks for similar words and synthetic data is generated according to distribution, so we are only interested in exact matches here. The output of the tool contains the index of the record in the original dataset and the index of the record in the synthetic dataset variant that are considered possible matches, as well as a score for each QI. These scores range from 0 to 1, where 1, indicates a possible match and 0, indicates not a possible match.

We filter the score results by applying a threshold of 0.5 for numeric Quasi-Identifiers and 1 for categorical attributes. In scenarios where potential matches exist across all QIs (worst-case scenario), we aggregate these results to determine the number of instances where each record in the original dataset corresponds to only one possible match in the synthetic variant.

\section{Results}\label{sec:results}
In this section, we present the results obtained from the development focusing on data utility and linkage attack.

\subsection{Data Utility}
We report the utility results for each synthetic data variant, as outlined in Table~\ref{tab:metrics}. The metrics, calculated for each attribute, are averaged across each variant.

\begin{figure}[h]
  \centering
  \includegraphics[width=0.75\linewidth]{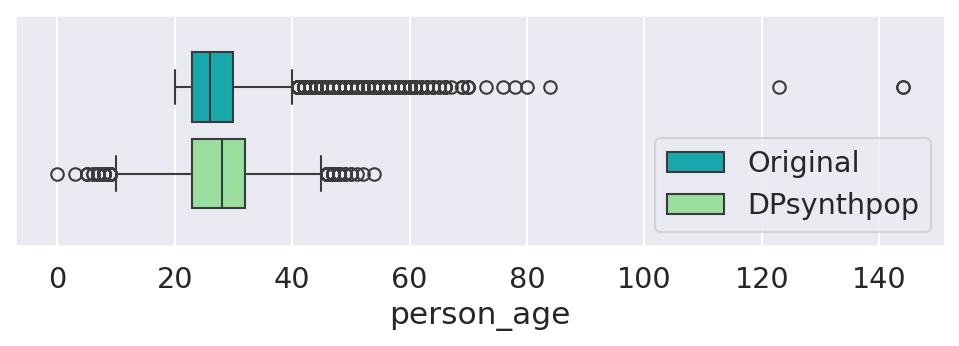}
  \caption{Example of the distribution of the attribute {\itshape person\_age} in the original dataset compared to a DPsynthpop variant.}
  \label{fig:comp1}
\end{figure}

Regarding BoundaryAdherence, all models scored a 1, indicating that data points were generated within established bounds. However, the DPsynthpop model performed poorly, with a median BoundaryAdherence of approximately 0.45, as it failed to adhere to the minimum and maximum values of the original dataset's attributes. Figure~\ref{fig:comp1} illustrates this issue by comparing the distributions of the original dataset and a specific DPsynthpop variant.

\begin{figure}[h]
  \centering
  \includegraphics[width=\linewidth]{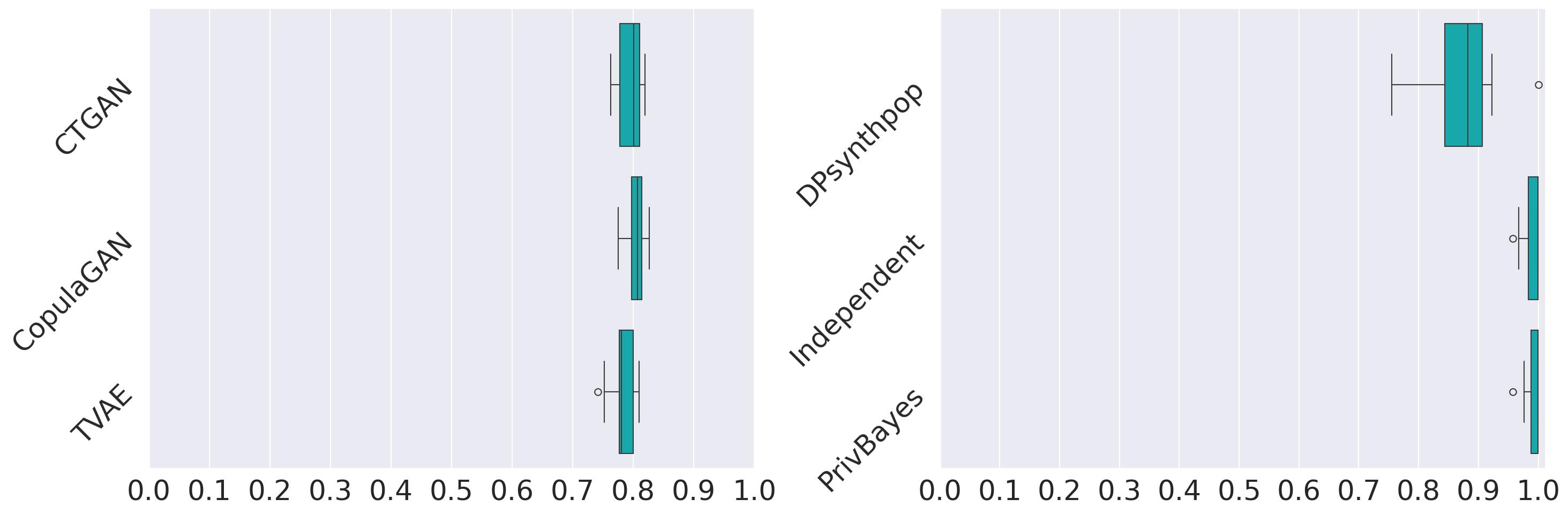}
  \caption{AttributeCoverage of the synthetic dataset variants generated with deep learning-based models (left) and differential privacy-based models (right).}
  \label{fig:coverage}
\end{figure}

Figure~\ref{fig:coverage} presents the AttributeCoverage for each synthetic data variant. Results show that differential privacy-based models outperform deep learning-based models, likely because the latter focuses on replicating the most frequent values from the original dataset. In contrast, differential privacy-based models, generate records to cover all possible values of each attribute, resulting in an overrepresentation of outliers compared to the original dataset (617 outliers). 
Table~\ref{tab:out} provides the average number of outliers generated by each model.

\begin{table}[ht!]
\centering
  \caption{Average number of outliers for each synthetic data generation model.}
  \label{tab:out}
  \scriptsize
  \begin{tabular}{lll}
    \toprule
    \textbf{Tool} & \textbf{Models} & \textbf{Outliers}\\
    \midrule
    & TVAE & 720\\
    SDV & CTGAN & 840\\
    & CopulaGAN & 794\\
    \hline
    & Independent & 1449\\
    DPART & PrivBayes & 1483\\
    & DPsynthpop & 114\\
  \bottomrule
\end{tabular}
\end{table}

\begin{figure}[h]
  \centering
  \includegraphics[width=\linewidth]{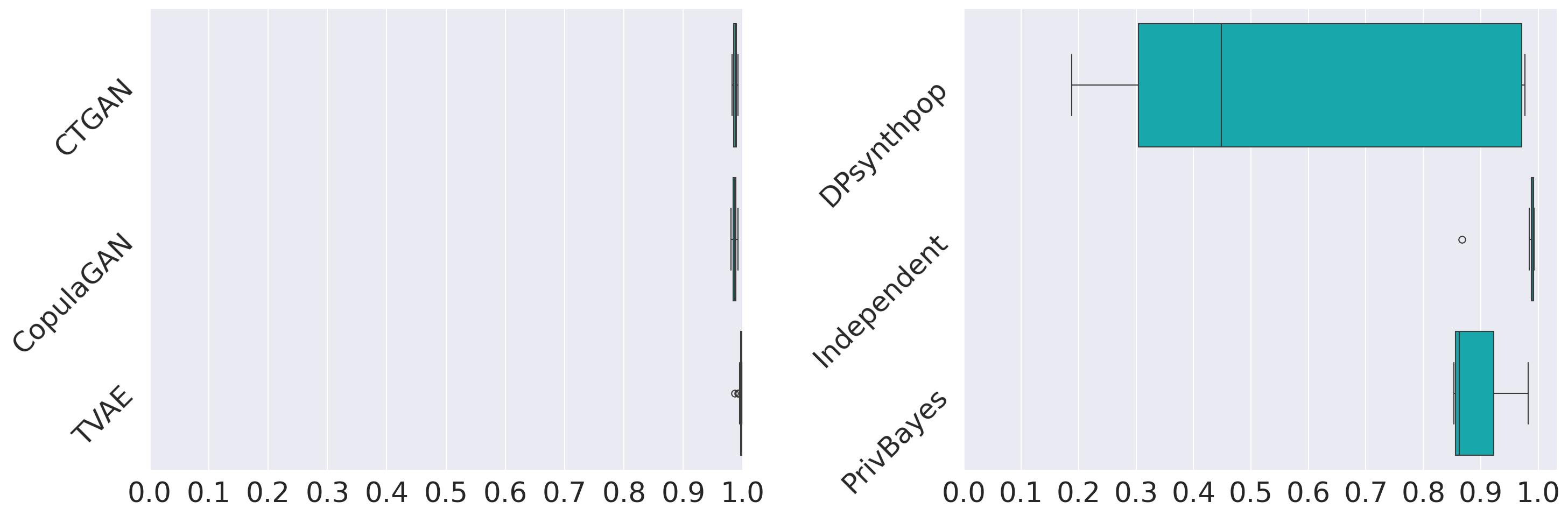}
  \caption{StatisticSimilarity of the synthetic dataset variants generated with deep learning-based models (left) and differential privacy-based models (right).}
  \label{fig:similarity}
\end{figure}

Note that DPsynthpop is an exception. As previously mentioned, this model fails to respect attribute value boundaries (Figures~\ref{fig:comp1} and~\ref{fig:coverage}), resulting in incomplete coverage of all possible attribute values. This model also demonstrates weak performance in StatisticSimilarity, with a median lower than 0.5 as shown in Figure~\ref{fig:similarity}. This indicates a poor replication of the original data distributions. All the remaining models show statistical similarity with the original data.

\subsection{Linkage Attack}
\begin{figure*}[ht!]
  \centering
  \includegraphics[width=\linewidth]{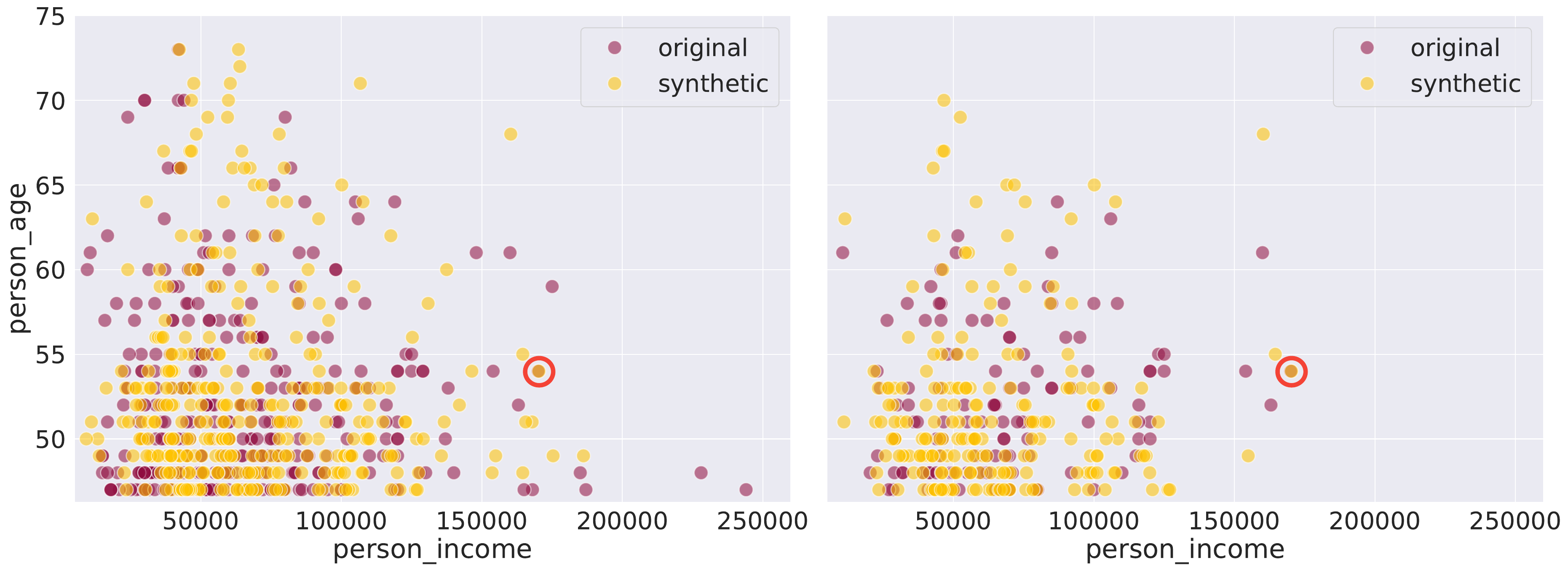}
  \caption{Distribution of records that are possible matches regarding the {\itshape person\_age} and {\itshape person\_income} attributes (left) and also including {\itshape person\_home\_ownership} and {\itshape loan\_intent} attributes (right) with one possible match highlighted.}
  \label{fig:res}
\end{figure*}

We now demonstrate the feasibility of re-identification using synthetic data through linkage attacks, particularly by leveraging outliers. We select a certain synthetic data variant, for demonstrative purposes. Specifically, we use a TVAE variant ({\itshape epochs} = 150, {\itshape batch\_size} = 20, {\itshape embedding\_dim} = 12). Our objective is to show the effectiveness of this attack by expanding the supposed background knowledge of an attacker. Figure~\ref{fig:res} presents on the left the potential matches based on the numerical QIs by comparing this variant against the original (4279 possible matches). In comparison, the right image shows the possible matches for all QIs (490 possible matches). For visual clarity, we only illustrate the potential matches for numerical QIs for individuals aged over 45.

In both cases, we observe multiple potential matches where the original and synthetic variant data points overlap. However, as more QIs are incorporated, the number of these potential matches decreases. This is due to the refined specificity of background knowledge, which narrows the pool of individuals who match these characteristics. Focusing on the highlighted example, Table~\ref{tab:recs} verifies these overlapping points. This particular match was obtained by filtering results for {\itshape person\_age} = 54 and 160000 <= {\itshape person\_income} <= 180000.

\begin{table}[ht!]
\centering
\scriptsize
  \caption{Example of possible matched records.}
  \label{tab:recs}
  \begin{tabular}{lll}
    \toprule
    \textbf{Attribute} & \textbf{Original} & \textbf{Variant}\\
    \midrule
    \textbf{{\itshape person\_age}} & \textbf{54} & \textbf{54}\\
    \textbf{{\itshape person\_income}} & \textbf{170000} & \textbf{170262}\\
    \textbf{{\itshape person\_home\_ownership}} & \textbf{MORTGAGE} & \textbf{MORTGAGE}\\
    {\itshape person\_emp\_length} & 12.0 & 33.0\\
    \textbf{{\itshape loan\_intent}} & \textbf{PERSONAL} & \textbf{PERSONAL}\\
    {\itshape loan\_grade} & A & A\\
    {\itshape loan\_amnt} & 11000 & 8053\\
    {\itshape loan\_int\_rate} & 6.62 & 7.27\\
    {\itshape loan\_status} & 0 & 0\\
    {\itshape loan\_percent\_income} & 0.06 & 0.06\\
    {\itshape cb\_person\_default\_on\_file} & N & N\\
    {\itshape cb\_person\_cred\_hist\_length} & 20 & 21\\
  \bottomrule
\end{tabular}
\end{table}

\begin{figure}[ht!]
  \centering
  \includegraphics[width=\linewidth]{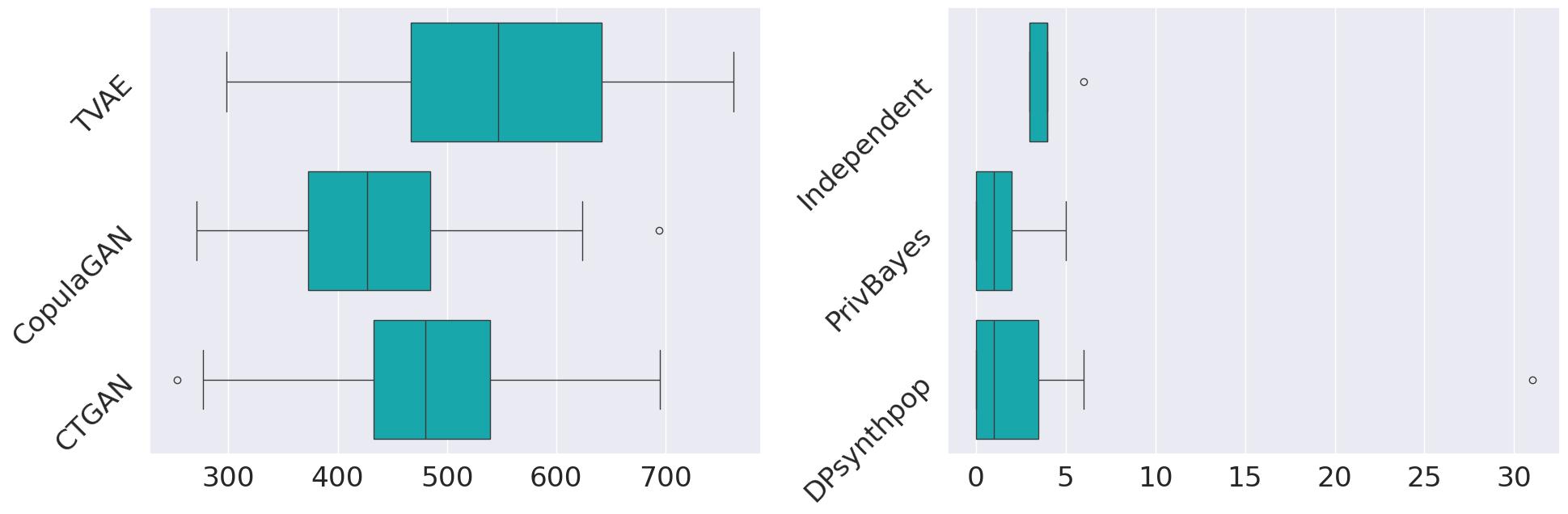}
  \caption{Possible matches for the synthetic dataset variants generated with deep learning-based models (left) and differential privacy-based models (right).}
  \label{fig:pm}
\end{figure}

Despite some attribute values being distant, all QIs matched. Given that these cases are outliers, and the models typically do not generate outliers considering non-outlier records, it is very unlikely that another data point could be a possible match. Therefore, we can infer that this variant record is a synthetic version of that original record, illustrating the feasibility of re-identifying synthetic data. Figure~\ref{fig:pm} shows the number of possible matches for each model’s synthetic dataset variants under the worst-case scenario -- using all QIs.

Deep learning-based models yield a higher number of possible matches compared to differential privacy-based models, indicating a heightened risk of re-identification. Although some of these values may be false positives due to the adjustments of scale and offset to widen our search of possible matches, the potential for privacy breaches remains substantial. Nevertheless, we considered the scores obtained on the attributes of the possible matches to assess success rate.

\section{Discussion}\label{sec:discussion}
Our experiments demonstrate that the protection of outliers ultimately depends on the synthetic data generation model. Deep learning-based models tend to adhere to the general distribution of the original dataset, focusing on generating more frequent values. On the other hand, differential privacy-based models deliberately generate records that cover all the values of each attribute, which results in a disproportionately high number of outliers (Table~\ref{tab:out}).

\begin{figure}[ht!]
    \begin{minipage}[c]{0.48\linewidth}
        \centering
        \includegraphics[width=\linewidth]{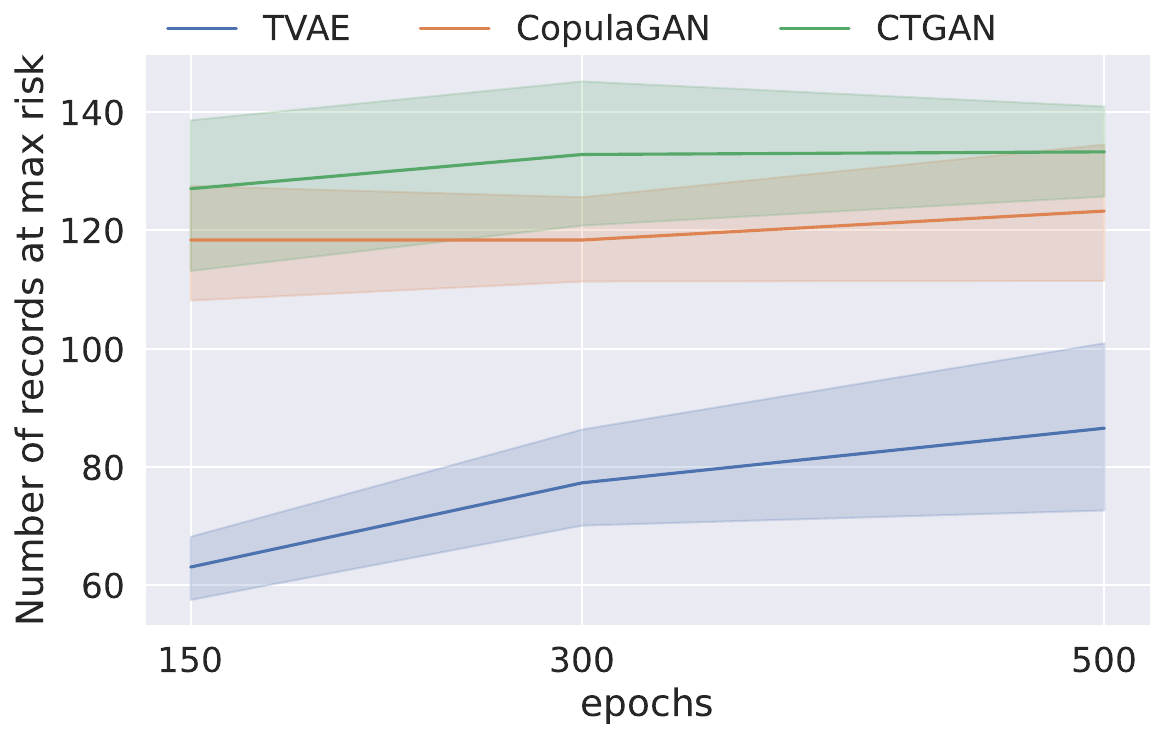}
        \caption{Original data records with one possible match in each synthetic data variant generated from deep learning-based models w.r.t {\itshape epochs} parameter.}
        \label{fig:risk}
    \end{minipage}\hfill
    \begin{minipage}[c]{0.48\linewidth}
        \centering
        \includegraphics[width=\linewidth]{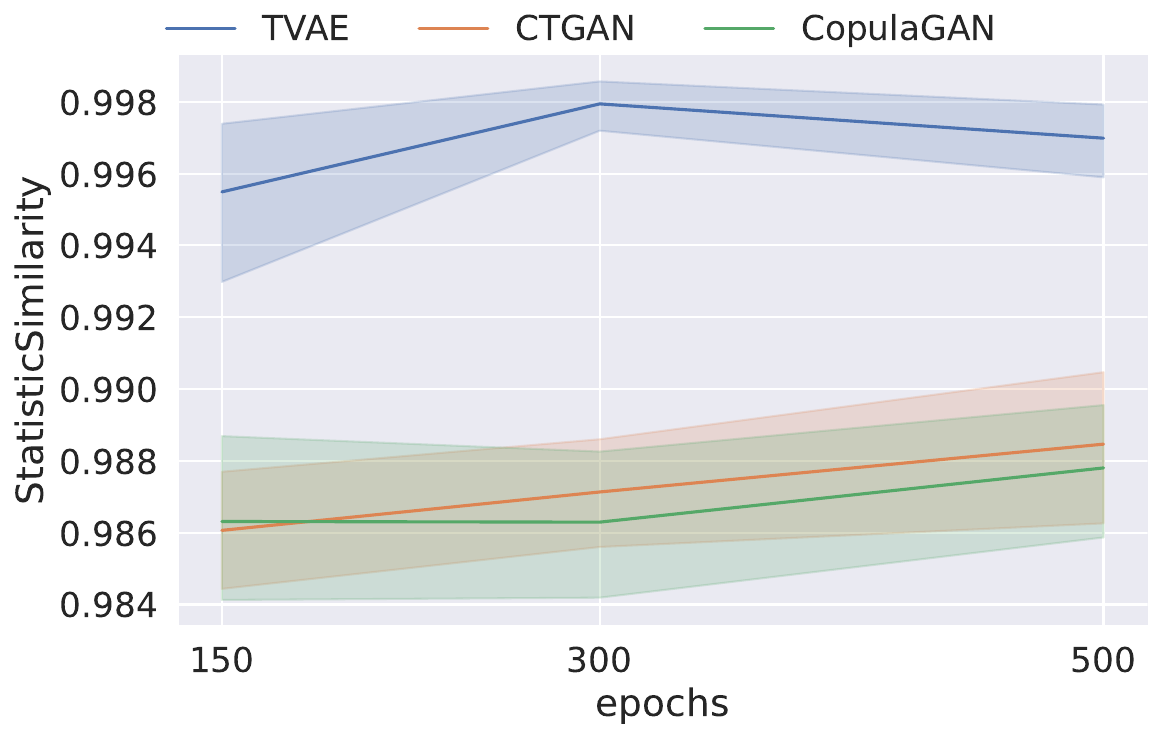}
        \caption{StatisticSimilarity of each synthetic data variant generated from deep learning-based models w.r.t {\itshape epochs} parameter.}
        \label{fig:util}
    \end{minipage}
\end{figure}

We also observe that adding an extra layer of protection, namely differential privacy, compromises data quality. Notably, the DPsynthpop model generated inferior quality synthetic data because it failed to accurately learn the distribution of the original dataset (Figures~\ref{fig:coverage} and~\ref{fig:similarity}). Despite this, the extra layer of protection resulted in fewer instances where only one possible match existed between each record of the original dataset and the synthetic variants. However, this enhancement in privacy protection might not outweigh the substantial loss in data quality and utility. To further understand this relationship, we present in Figure~\ref{fig:risk} and Figure~\ref{fig:util} the number of cases that are at maximum risk (1 possible match) for each synthetic data variant concerning the {\itshape epochs} parameter, and how similar each synthetic data variant is concerning this parameter. We do not include differential privacy-based models in this analysis as the parameters are not comparable and they perform very poorly in terms of data utility. This leads us to the question: {\itshape how does the number of epochs influence utility and privacy?}

Results indicate that as the parameter {\itshape epochs}, which represents the number of iterations the models use to optimize their parameters, increases, the number of unique matches between the original dataset and the synthetic variants also rises. For variants generated using CTGAN and CopulaGAN, increased {\itshape epochs} enhance the similarity between the original and synthetic data, whereas for those generated using TVAE, the similarity decreases. In general, a higher {\itshape epoch} value allows models to better capture the characteristics of data, resulting in synthetic data that more closely resemble the original. Consequently, this similarity also elevates the re-identification risk.  It is therefore important to tune the hyperparameters of synthetic data generation models according to the data's characteristics to achieve an optimal balance between data quality and privacy.

For future work, we plan to include records that are not outliers, a larger number of datasets, different QI sets, and other types of attacks to corroborate these results. We considered the risk of MIA using the DOMIAS~\cite{van2023membership} tool which, to the best of our knowledge, is the only tool performing MIA on synthetic data. DOMIAS is a density-based MIA model that allows MIA against synthetic data by targeting local overfitting of the generative model. Unfortunately, a big limitation of this tool is that it only operates on numerical data. Furthermore, we aim to investigate the impact of outlier treatment strategies at different stages (before, during, and after synthesis) on both utility and privacy. Also, evaluate the necessity of excluding outliers from the synthetic data generation process.

Although synthetic data does not fully protect the privacy of individuals, it serves as an effective proxy for the original data in many tasks, highlighting its value. We stress the importance of developing and updating synthetic generation approaches for a secure and robust data environment that balances utility with privacy considerations specially focused on extreme and rare data points.

\section{Conclusion}\label{sec:conclusion}
In this paper, we conducted an analysis focused on 
the effectiveness of the re-identification associated with synthetic data generation models, specifically examining their efficacy in protecting extreme data points, i.e. outliers.

Our results showed that outlier protection is model-dependent. The deep learning-based models tested focused on generating more frequent values, while the differential privacy-based models generally generated a higher number of outliers. However, the differential privacy-based models also resulted in poorer data quality. Most importantly, we conducted a linkage attack to demonstrate how outliers can be exploited to re-identify personal information, highlighting the vulnerability associated with synthetic data.

\begin{credits}

\subsubsection{\discintname}
The authors have no competing interests to declare that are relevant to the content of this article.
\end{credits}
%
%
%
\bibliographystyle{splncs04}
\bibliography{mybib}
%




\end{document}